\documentclass[letterpaper, 10 pt, journal, twoside]{IEEEtran}

\IEEEoverridecommandlockouts                              

\usepackage{hyperref}
\usepackage{cite}
\usepackage{graphicx}

\usepackage{subcaption}

\usepackage{booktabs}

\usepackage{amsmath,amssymb,amsfonts}
\usepackage{algorithmic}
\usepackage{textcomp}
\usepackage{bm}
\graphicspath{./pics/}
\usepackage[final]{changes}   
\usepackage{makecell}

\usepackage{eso-pic}

\DeclareGraphicsExtensions{.pdf,.jpeg,.png,.jpg}

\begin{document}

\title{\LARGE \bf
Multi-Touch and Bending Sensing Using Electrical Impedance Tomography for Robotics}

\author{Haofeng Chen$^{1}$, Bedrich Himmel$^{1}$, Bin Li$^{2}$,  Xiaojie Wang$^{\ast,2}$, Matej Hoffmann$^{\ast,1}$
\thanks{Manuscript received: September 11, 2025; Revised: February 27, 2026; Accepted: June 3, 2026.}
\thanks{This paper was recommended for publication by Editor Yong-Lae Park upon evaluation of the Associate Editor and Reviewers comments.}
\thanks{This work was co-funded by the European Union under the project Robotics and Advanced Industrial Production (reg. no. CZ.02.01.01/00/22\_008/0004590). B.L. and X.W. were supported by the Frontier Technologies R\&D Program of Jiangsu (Grant No. BF2025057). B.H. and M.H. were additionally supported by the Czech Science Foundation (GA ČR), project no. 25-18113S.}
\thanks{$^{1}$Haofeng Chen, Bedrich Himmel, Matej Hoffmann are with the Department of Cybernetics, Faculty of Electrical Engineering, Czech Technical University in Prague  {\tt\footnotesize matej.hoffmann@fel.cvut.cz}}%
\thanks{$^{2}$Bin Li and Xiaojie Wang are with the Institute of Intelligent Machines, Hefei Institutes of Physical Science, Chinese Academy of Sciences, Hefei 230031, China  {\tt\footnotesize xjwang@iamt.ac.cn}}
\thanks{$\ast$ Shared corresponding authors.}
\thanks{Digital Object Identifier (DOI): see top of this page.}
}


\AddToShipoutPictureFG*{%
  \AtPageUpperLeft{%
    \raisebox{-1.1cm}{
      \makebox[\paperwidth]{%
        \begin{minipage}{0.92\paperwidth}
          \centering
          \color{gray}

          {\fontsize{7}{8}\selectfont
          This is the authors' final version of the manuscript published as:\\
          Chen, H.; Himmel, B.; Li, B.; Wang, X. \& Hoffmann, M. (2026), 
          'Multi-Touch and Bending Sensing Using Electrical Impedance Tomography for Robotics', \\IEEE Robotics and Automation Letters 11(8), 9851-9858, \textcopyright~IEEE,
          \texttt{https://doi.org/10.1109/LRA.2026.3709063}\\
         \par}
        \end{minipage}%
      }%
    }%
  }%
}

\maketitle

\begin{abstract}

Electrical Impedance Tomography (EIT) offers a promising solution for distributed tactile sensing with minimal wiring and full-surface coverage in robotic applications. However, EIT-based tactile sensors face significant challenges during surface bending. Deformation alters the baseline impedance distribution and couples with touch-induced conductivity variations, complicating signal interpretation. To address this challenge, we present a novel sensing framework that integrates a deep neural network for interaction state classification with a dynamic adaptive reference strategy to decouple touch and deformation signals, while a data-driven regression model translates EIT voltage changes into continuous bending angles. The framework is validated using a magnetic hydrogel composite sensor that conforms to bendable surfaces. Experimental evaluations demonstrate that the proposed framework achieves precise and robust bending angle estimation, high accuracy in distinguishing touch, bending, and idle states, and significantly improves touch localization quality under bending deformation compared to conventional fixed-reference methods. Real-time experiments confirm the system’s capability to reliably detect multi-touch interactions and track bending angles across varying deformation conditions. This work paves the way for flexible EIT-based robotic skins capable of rich multimodal sensing in robotics and human–robot interaction.

\end{abstract}

\section{Introduction}
The importance of endowing robots with touch has been recognized for several decades and a large number of technologies have been developed (e.g., \cite{Bartolozzi2016,Dahiya2019} for surveys). The focus has largely been on tactile sensing for manipulation, as equipping robot fingers or hands with tactile sensors requires only relatively small patches of electronic skin. Whole-body artificial skins for robots have been an exception, with only few successful solutions such as \cite{Cannata2008,Mittendorfer2011a}. The potential of sensing over the whole body surface has also not been fully explored. Cheng et al. provide an overview of applications of their electronic skin in \cite{Cheng2019comprehensive}.

Even for the so-called whole-body skins \cite{Cheng2019comprehensive,Maiolino2013}, it is not the complete surface of the robot that is covered. In particular the areas around the robot joints are typically without skin coverage. Complete coverage would be desired for both physical and social human-robot interaction \cite{RN310,RN444}. The main challenge for articulated robots lies in sustaining large deformation and stretch while still delivering touch information. While measuring the angle of deformation (bending) may not be key for classical robots, which have accurate joint encoders, bending information may be relevant in soft robots, providing valuable information about their body configuration (see e.g., \cite{thuruthel2019soft}).
\begin{figure}[t]
	\centering
	\includegraphics[scale=0.88]{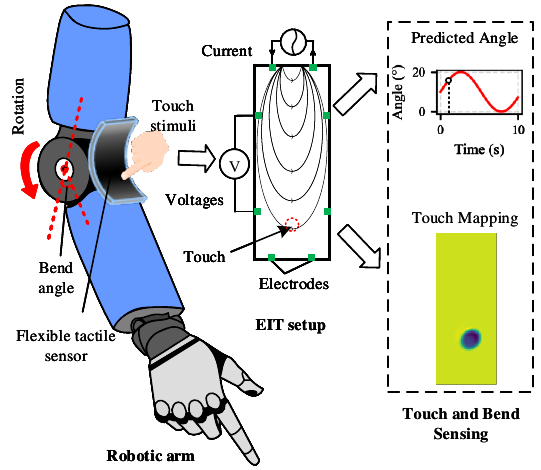}
	\caption{Overview of EIT-based flexible tactile sensor for touch and bending sensing.} \label{diagram_concept}
\end{figure}

While flexible tactile sensors have been explored, few achieve both high-resolution touch and bending measurement.
For example, Noda et al. \cite{RN1163} introduced a stretchable liquid-based tactile sensor with a dual-channel design to detect bending and touch, but its low spatial resolution limits multi-touch sensing. Totaro et al.~\cite{RN1159} proposed a multimodal mechanosensing system using resistive and optical components, capable of distinguishing between pressure and bending, but limited to single-point touch and complex in design due to multi-principle integration. Sarwar et al.~\cite{sarwar2017bend} developed a capacitive sensor array capable of detecting multi-touch during bending and stretching, but it does not explicitly decouple touch and deformation signals and lacks quantitative bending estimation. Zhang et al.~\cite{RN1161} designed a microwave-based sensor capable of sensing both pressure and bending, but its small sensing area and electromagnetic coupling mechanism limit scalability for large-area applications. Overall, most sensors capable of touch and bending measurement compromise between spatial resolution, multi-touch capability, and signal decoupling, making it difficult to achieve both high‑resolution touch and bending measurement. As the sensing area increases, the growing number of sensors adds wiring and interface complexity, along with higher costs.

\begin{table}[t]
\centering
\caption{Comparison of flexible tactile sensors supporting touch or bending sensing.}
\scriptsize
\setlength{\tabcolsep}{2pt}
\begin{tabular}{lccccc}
\toprule
\textbf{Study} & \textbf{Method} & \textbf{Touch\&} & \textbf{Multi-} & \textbf{Area} & \textbf{Flexible} \\
 &  & \textbf{Bend} & \textbf{Touch} &  \\
\midrule
Sarwar et al.~\cite{sarwar2017bend} & Capacitive & Yes & Yes & 55$\times$55 $mm^2$ & Yes \\
Kim et al.~\cite{RN1160}            & \makecell{Optical+\\Piezo-resistive} & Yes & No  & 70$\times$10 $mm^2$ & Yes \\
Lee et al.~\cite{RN1158}            & Resistive     & Bend & No & 10$\times$10 $mm^2$ & Yes \\
Xin et al.~\cite{RN711}             & EIT           & Bend & No & 60$\times$20 $mm^2$ & Yes \\
Zhang et al.~\cite{RN1161}          & Microwave     & Yes  & No & 95$\times$58 $mm^2$ & Yes \\
Yu et al.~\cite{RN720}              & EIT           & Yes  & No & $\sim$952 $cm^2$ & Yes\\
Dong et al.~\cite{RN1176}           & EIT+3D Scan   & Touch & Yes & 150$\times$100 $mm^2$ & Yes \\
Park et al.~\cite{RN1153}           & EIT           & Touch & Yes & 314 $mm^2$ & No \\
\textbf{This Work}                  & \textbf{EIT}  & \textbf{Yes} & \textbf{Yes} & \textbf{150$\times$60 $mm^2$} & Yes\\
\bottomrule
\end{tabular}
\label{tab:comparison}
\end{table}

EIT presents a promising alternative for distributed sensing without requiring multiple discrete sensor units, making it well suited for whole-body robotic applications. Prior research has explored EIT for touch sensing, such as Kato et al.~\cite{RN1016}, who developed a soft, pressure-sensitive EIT-based tactile sensor for finger touch sensing. Nagakubo et al. \cite{RN1029} presented an early demonstration of EIT-based tactile sensing by attaching a conductive knit sensor to a dummy’s elbow. They applied pressure to one point under fully bent and fully extended postures, showing that EIT can detect touch on a deformable surface. However, their work focused on basic touch detection and did not explore continuous bending sensing or the interaction between touch and bending. Yu et al. \cite{RN720} developed uKnit, a scarf-like soft wearable sensor combining machine knitting with EIT. The system demonstrated worn-location detection, gesture recognition, and respiratory monitoring, highlighting EIT’s potential for flexible and reconfigurable sensing. However, its design faced challenges such as limited spatial resolution and signal anisotropy due to the knitted fabric structure, which constrains sensing precision.
Park et al.\cite{RN1153} introduced a large-area, face-shaped EIT sensor for multi-touch recognition. 

EIT has also been applied to shape sensing in soft robots, with Avery et al. \cite{RN408} and Xin et al. \cite{RN711} using it to map deformations in real time. However, these approaches lack touch sensing, limiting their practicality for real-time applications. 
Dong et al. \cite{RN1176} demonstrated 3D-scanner-assisted EIT for tactile interaction mapping on deformable surfaces, but their approach required complex hardware and was not suitable for real-time applications. Despite these advancements, few studies have addressed the sensing of touch and bend in large-deformation robotic applications using EIT. 
Table \ref{tab:comparison} summarizes recent developments in flexible tactile sensors capable of touch and/or bend sensing.  

The primary challenge in achieving both touch and bending sensing using EIT lies in signal coupling---both touch and bending alter the impedance distribution in a complex manner, making it difficult to separate their effects. Additionally, EIT is an ill-posed inverse problem, where small measurement variations can lead to significant reconstruction errors, especially under substantial deformations.

To address these limitations, this paper presents an EIT-based sensing framework capable of both real-time multi-touch detection and bending angle estimation, with enhanced touch resolution even after significant deformation of the sensor surface. Figure \ref{diagram_concept} is an illustration of  the flexible tactile sensor on a robot joint, enabling it to track bending angles and localize touch positions through conductivity distribution analysis.  EIT reconstructs internal conductivity patterns by applying electrical currents and measuring boundary voltages via surface electrodes. Touch interactions generate localized conductivity mapping, whereas bending leads to distributed measurement variations due to geometric deformation. The deformation information can be leveraged to enhance touch localization and enable accurate bending estimation.

The key contributions of this work are summarized as follows:  
(1) We propose a novel sensing framework based on EIT that enables both localization of multi-touch events and estimation of bending angles on a single flexible sensor surface.  

(2) To address the signal coupling between touch and bending, we introduce a dynamic adaptive reference strategy that updates reconstruction references based on the deformation state, enhancing touch localization resolution under large surface deformations. 

(3) We design and implement a large-area, conformable EIT sensor suitable for robotic integration, capable of maintaining high spatial resolution across different bending configurations.  

(4) We demonstrate real-time operation of the proposed system in experiments involving sequential bending and touch inputs, highlighting its potential for interactive robotic applications.

\section{METHODS AND SYSTEM DESIGN}
\begin{figure*}[htbp] 
\centering 
\includegraphics[width=0.92\textwidth]{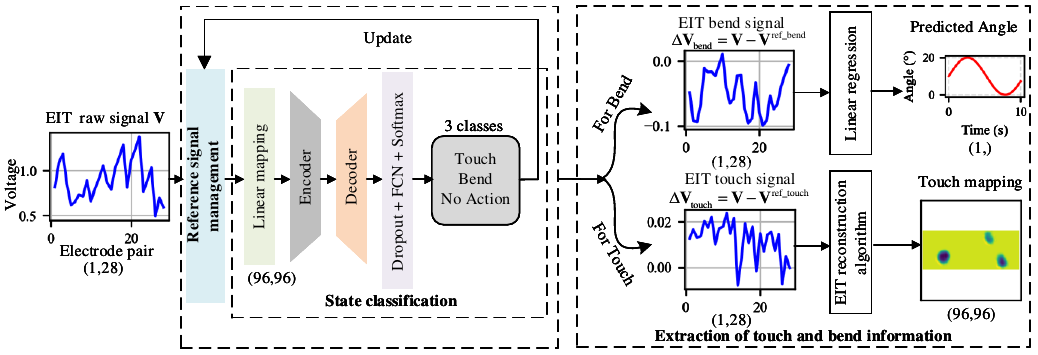} 
\caption{Overview of the proposed framework for multi-touch and bending sensing.} 
\label{network_all} 
\end{figure*}

The proposed method enables sensing of touch and bend based on EIT signals. As shown in Figure \ref{network_all}, the framework comprises three main stages: state classification, reference signal management, and extraction of touch and bend information. Our system is built around time-difference EIT modeling, which estimates conductivity changes $\Delta \boldsymbol{\sigma}$ from changes in boundary voltage $\Delta \mathbf{V}$ across temporal states. This model serves as a foundational transformation across all processing stages, including feature extraction, classification, and touch mapping.

\subsection{Time-Difference EIT Modeling}

The EIT problem involves reconstructing the internal conductivity distribution of a conductive domain based on applied currents and measured boundary voltages. We use a time-difference formulation \cite{RN309} to emphasize temporal changes in conductivity between a reference and observed state.
In difference imaging, the connection between the change in conductivity, $\Delta \boldsymbol{\sigma} \in {\mathbb{R}^N}$, and the change in measured voltage, 
$\Delta \mathbf{V} \in {\mathbb{R}^M}$
, can be approximated as a linear relationship:

\begin{equation}
\Delta \mathbf{V} = {\text{ }}\mathbf{J}\Delta \boldsymbol{\sigma} {\text{  +  }}{\mathbf{n}_{noise}}
\label{eq1_dEIT}
\end{equation} 
where $\mathbf{J}{\text{ }} \in {\mathbb{R}^{M \times N}}{\text{ }}(M \ll N)$ is the sensitivity (Jacobian) matrix, calculated via the finite element method (FEM), ${\mathbf{n}_{noise}} \in {\mathbb{R}^M}$ accounts for measurement noise.
We define the conductivity change vector as $\Delta \boldsymbol{\sigma} {\text{  =  }}\left[ {\Delta {\sigma _1},\Delta {\sigma _2},...,\Delta {\sigma _N}} \right]$, where $\Delta {\sigma _i} = {\text{ }}{\sigma _i} - {\sigma _0}$. Here, ${\sigma _0}$ is the background conductivity, and ${\sigma _i}$ is the conductivity of the $i$-th element. Similarly, the voltage change vector 
$\Delta \mathbf{V}{\text{  =  }}\left[ {\Delta {v_1},\Delta {v_2},...,\Delta {v_M}} \right]$ is calculated as 
$\Delta {v_i}{\text{  =  }}{v_i}{\text{ }} - {\text{ }}v_i^{ref}$, where 
${\mathbf{V}^{ref}}{\text{ =  }}\left[ {v_1^{ref},v_{_2}^{ref},...,v_M^{ref}} \right]$ is the reference voltage vector and 
$\mathbf{V}{\text{  =  }}\left[ {{v_1},{v_2},...,{v_M}} \right]$ is the current voltage vector. 
To solve the inverse problem, we employ the one-step Gauss-Newton method \cite{RN656}: 
\begin{equation}
\Delta \boldsymbol{\sigma} {\text{ }} = {\text{ }}{({\mathbf{J}^T}\mathbf{J} + {\lambda ^2}\mathbf{R})^{ - 1}}{\mathbf{J}^T}\Delta \mathbf{V}
\label{eq1}
\end{equation}
where $\mathbf{J}^T$ represents the transpose of the sensitivity matrix $\mathbf{J}$, $\lambda$ is a scalar hyperparameter that controls the regularization strength, chosen heuristically.
$\mathbf{R}$ denotes the regularization matrix. The matrix 
$\mathbf{R}$ is computed using Newton’s One-Step Error Reconstructor (NOSER) prior\cite{RN983}. 
This transformation is treated as a linear mapping in our neural pipeline and is also used for touch reconstruction after classification.

\subsection{Reference signal management and state classification}

In dynamic tactile applications, the sensor undergoes significant deformation, which affects the reference voltage readings. To ensure accurate differential imaging, we adopt a state-aware classification framework that also manages reference signal updates adaptively.

The pipeline begins with the sensor’s raw voltage signal $\mathbf{V} \in \mathbb{R}^{1 \times 28}$, which is first processed by a reference signal management module (Figure \ref{network_all}). This module manages two types of reference voltages. The first is a static reference, denoted as $\mathbf{V}^{\text{ref}\_{\text{bend}}}
$, acquired when the sensor is in its undeformed state and used as the reference for bending estimation. The second is a dynamic touch reference, denoted as $\mathbf{V}^{\text{ref}\_{\text{touch}}}$, which is recorded after deformation when a touch event is detected under bent conditions. For classification, the system compares the current voltage measurement with the static reference to obtain a differential signal that reflects internal conductivity changes. 

This difference signal is then linearly projected into an image-like conductivity embedding $\Delta \boldsymbol{\sigma}$ via Eq.~(\ref{eq1}). This transformation provides spatially meaningful features and reduces sensitivity to static boundary effects, allowing the classifier to focus on internal conductivity variations caused by touch or bending. 
The conductivity map is then fed into an encoder–decoder neural network for interaction classification. The encoder consists of three convolutional layers with kernel size 3×3 and progressively increasing feature dimensions from 16 to 32 and then to 64, interleaved with max-pooling and batch normalization layers. The decoder mirrors this structure using transposed convolutions and upsampling operations to restore spatial resolution. Finally, two fully connected layers reduce the representation to 16 dimensions, followed by a softmax layer that predicts the current sensor state as either touch, bend, or no action.

The classification result guides the reference management process. When a bending state is detected, the system retains the static reference for subsequent angle estimation. When a touch is detected under deformation, the dynamic reference is updated to enable accurate touch reconstruction. These references are then used in the following stage for bend angle prediction and touch mapping.

\subsection{Bend angle prediction and Touch mapping}
Task-specific EIT signals are extracted using bend reference for bending and touch reference for touch, forming the basis for subsequent modeling (Figure \ref{network_all}). To establish a quantitative relationship between sensor voltage changes and bending angles, we adopted a data-driven approach combining feature selection and linear regression modeling from the scikit-learn library\cite{pedregosa2011scikit}.  Given the high-dimensional nature of the EIT voltage response data, we performed feature selection using SelectKBest with $f\_regression$ scoring from Scikit-learn\cite{pedregosa2011scikit}, leveraging the continuous nature of the bending angles. This process identified the most relevant voltage features based on their correlation with bending variations, effectively reducing dimensionality and focusing on the most informative signals. These selected features were then used to train a linear regression model, establishing a direct mapping between voltage changes and bending angles.

For touch mapping, the system employs the previously described one-step Gauss-Newton reconstruction algorithm to recover the conductivity distribution. To improve the spatial resolution and suppress reconstruction artifacts, a lightweight convolutional post-processing module is applied \cite{RN923}. This module enhances the clarity of contact regions without significantly increasing computational cost.

\subsection{Tactile sensor fabrication and hardware setup}

\begin{figure}[htbp] 
\centering 
\includegraphics[width=0.44\textwidth]{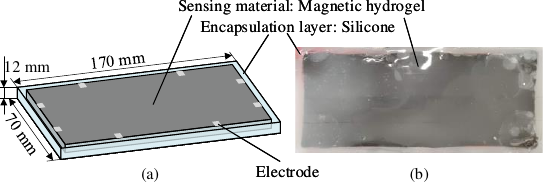} 
\caption{Illustration of an 8-Electrode EIT Tactile Sensor. (a) Schematic of the magnetic hydrogel-based EIT tactile sensor. (b) Photograph of the sensor.} 
\label{Fig4} 
\end{figure}
The magnetic hydrogel composite was chosen as the sensing layer in this study mainly because of its improved mechanical stability and strong pressure-sensing performance when integrated with EIT~\cite{RN1143}.
As shown in Fig. \ref{Fig4}, the tactile sensor is composed of three primary components: a silicone layer, a magnetic hydrogel sensing layer, and eight silver electrodes uniformly distributed along the boundaries. The sensor has overall dimensions of 160 mm × 70 mm × 12 mm, while the magnetic hydrogel layer measures 150 mm × 60 mm × 5 mm. Eight electrodes were employed as a balance between spatial resolution, hardware simplicity, and response time. Each of the 8 silver electrodes interfaces with the conductive layer through a contact area of 5 mm × 5 mm. To prevent moisture loss, the sensing material is encapsulated using Ecoflex 30 silicone rubber, which helps maintain the hydration state of the hydrogel and ensures the long-term stability of its performance. Components A and B of the silicone were mixed at a 1:1 weight ratio, stirred thoroughly, and degassed to eliminate air bubbles. The mixture was then cast into a mold, fully encapsulating the magnetic hydrogel. After curing at room temperature for 2–3 hours, a protective silicone layer was formed, ensuring the hydrogel’s stability and functionality.

The data acquisition circuit is based on the AD5940 impedance measurement IC, which supports signal generation and measurement bandwidth up to 200 kHz~\cite{RN277}. The AD5940 performs an on-chip Discrete Fourier Transform to obtain magnitude and phase values, while additional digital filtering (sinc and averaging filters) suppresses mains interference and improves the signal-to-noise ratio. A 40 kHz excitation frequency balances speed and accuracy, consistent with prior EIT systems~\cite{RN725, RN255}. Multiplexers (ADG706) enable sequential addressing of electrodes, and a 4-wire measurement approach captures impedance changes~\cite{RN277}, acquiring unique pairwise measurements without repetition. A microcontroller (STM32) coordinates the multiplexers, the AD5940, and data transmission to a computer. Further implementation details are provided in~\cite{RN906}.

\section{Tactile sensor experiment}
\begin{figure}[htbp] 
\centering 

\includegraphics[width=0.30\textwidth]{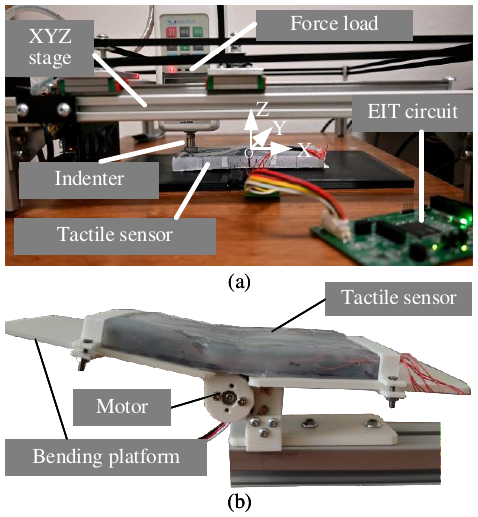} 
\caption{Experimental setups for (a) Touch test; (b) Bending test.} 
\label{Fig6} 
\end{figure}
Experiments were conducted to evaluate the touch sensing and bending angle estimation capabilities of the proposed EIT-based tactile sensor. As shown in Figure \ref{Fig6} (a), a controlled indentation test was performed using an experimental setup where the sensor was fixed horizontally, and an indenter was mounted on a computer-controlled XYZ stage. The stage was programmed to move along the z-axis to apply indentation forces, which were precisely measured using an inline commercial force gauge (Sauter FH100). To evaluate the sensor’s response to mechanical deformation, a bending test platform was developed, as shown in Figure \ref{Fig6} (b). The setup consists of a 3D-printed support structure with a mounted servo motor (ST3025, Waveshare Electronics, China) that drives a hinged mechanism to induce controlled bending of the tactile sensor. The sensor is fixed at both ends, and the central region is deflected by rotating the motor to predefined angles. The bending platform ensures repeatable and adjustable deformation, enabling systematic collection of EIT data under various bending conditions. The bending angle is precisely controlled via the motor, allowing synchronization between mechanical input and impedance measurement for training and validation purposes.
\subsection{Single-touch indentation test}

\begin{figure}[htbp] 
\centering 
\includegraphics[width=0.38\textwidth]{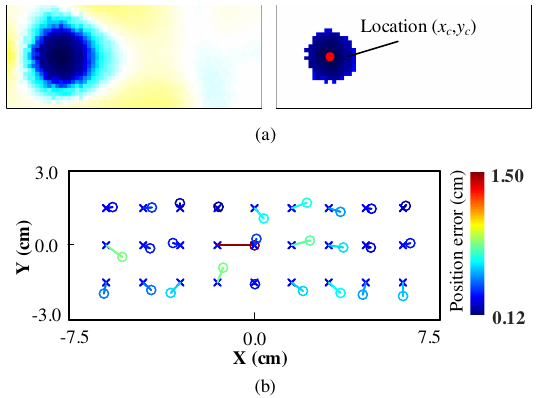} 
\caption{Touch position error results.} 
\label{PE_DL} 
\end{figure}
The sensor's touch localization accuracy was evaluated by measuring the error between the reconstructed compressed area's centroid and the actual indenter position. A 15 mm diameter indenter applied a constant 10 N force at 27 different positions, spaced 15 mm apart along both the x- and y-axes. A force of approximately 5 N is required to be reliably detected across the sensing area. The touch location was determined using the method described in \cite{RN906}. The touch positions are estimated by computing the centroid $(x_c, y_c)$ of the reconstructed image, as shown in Figure \ref{PE_DL} (a). Specifically, for the post-processed output, the centroid is calculated using a weighted average of pixel coordinates, where the weights correspond to the pixel intensities. This can be expressed as:
\begin{equation}
(x_c, y_c) = \left( \frac{\sum_i P_i x_i}{\sum_i P_i}, \; \frac{\sum_i P_i y_i}{\sum_i P_i} \right)
\label{center}
\end{equation}
Here, $P_i$ denotes the pixel value at coordinate $(x_i, y_i)$ in the post-processed image.
As shown in Figure \ref{PE_DL}(b), the error vectors represent the differences between the actual touch positions and the estimated locations. For each indentation point, we collected 8 repeated samples, resulting in an average localization error of 4.8 ± 2.8 mm.

\subsection{State classification test}
To classify different sensor interaction states (touch, bending, and idle), a supervised learning strategy was employed. The dataset consisted of 1,080 labeled samples, evenly distributed across the three classes to ensure balanced training. The bending class included 360 samples, each corresponding to a bending angle randomly selected within the range of $-20^\circ$ to $50^\circ$, which was chosen as a safe and stable operating window where the sensor maintains reliable electrode contact and measurement quality. The touch class comprised 360 samples captured from randomly chosen contact locations on the sensor surface, ranging from single-point to four-point touch. The idle class included 360 samples representing the no-action condition of the sensor.
The dataset was randomly partitioned into 60\% for training, 20\% for validation, and 20\% for testing. 
The deep neural network mentioned in the Figure \ref{network_all} was trained using the sparse categorical cross-entropy loss function, Adam optimizer with a learning rate of $1\times10^{-5}$, and a batch size of 64, over 150 epochs to optimize classification performance. Experiments were conducted on a PC with an Intel Core i7-10750H CPU and 16 GB RAM.

\begin{figure}[htbh]
	\centering
	\includegraphics[scale=0.86]{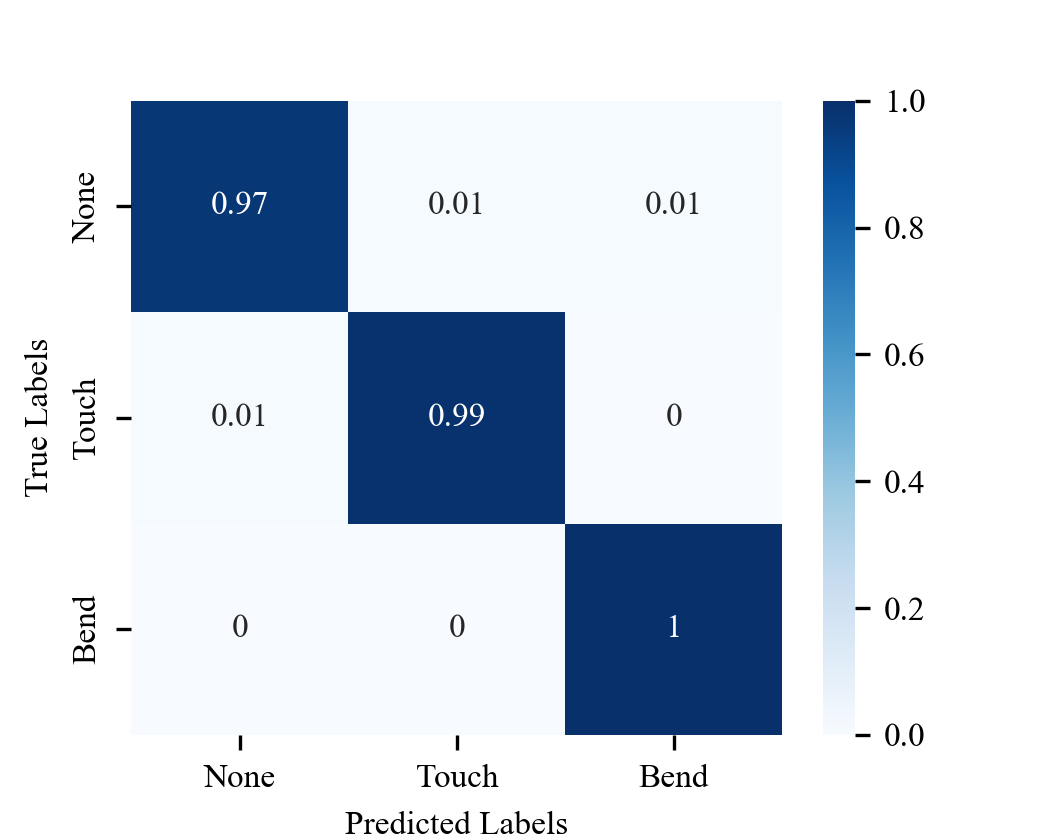}
	\caption{Normalized confusion matrix for touch, bending, and idle state classification on test dataset.} \label{Confusionmatrix}
\end{figure}

The classification performance on the test set is summarized in the normalized confusion matrix shown in Figure \ref{Confusionmatrix}. The model achieved high accuracy across all classes. This is attributed to the strong and distinctive features of bending-induced signals, which introduce significant static impedance changes that are easily distinguishable from other states.
In contrast, minor misclassifications occurred between the touch and idle (none) classes. Specifically, 1\% of the idle samples were misclassified as touch, and a similar proportion of touch samples were incorrectly labeled as idle. This confusion likely stems from the ill-posed nature of EIT reconstruction, where system noise and reconstruction artifacts—especially under low-contrast conditions—can produce localized patterns resembling touch inputs. This makes weak touch signals difficult to distinguish from background noise in the idle state.
Despite this, the classifier still achieved over 97\% accuracy for the idle class and 99\% for the touch class, demonstrating robust performance across varying sensor interaction conditions.

\subsection{Bending test}
To establish a quantitative relationship between sensor voltage changes and bending angles, we adopted a data-driven approach that combines feature selection with linear regression modeling from the scikit-learn \cite{pedregosa2011scikit}.  The model was trained using 360 bending samples, where the input features correspond to the EIT voltage change, and the output labels represent the ground-truth bending angles in degrees. The performance of this approach was evaluated through comprehensive testing of real-time angle prediction capabilities.

\begin{figure}[htbh]
	\centering
	\includegraphics[scale=0.80]{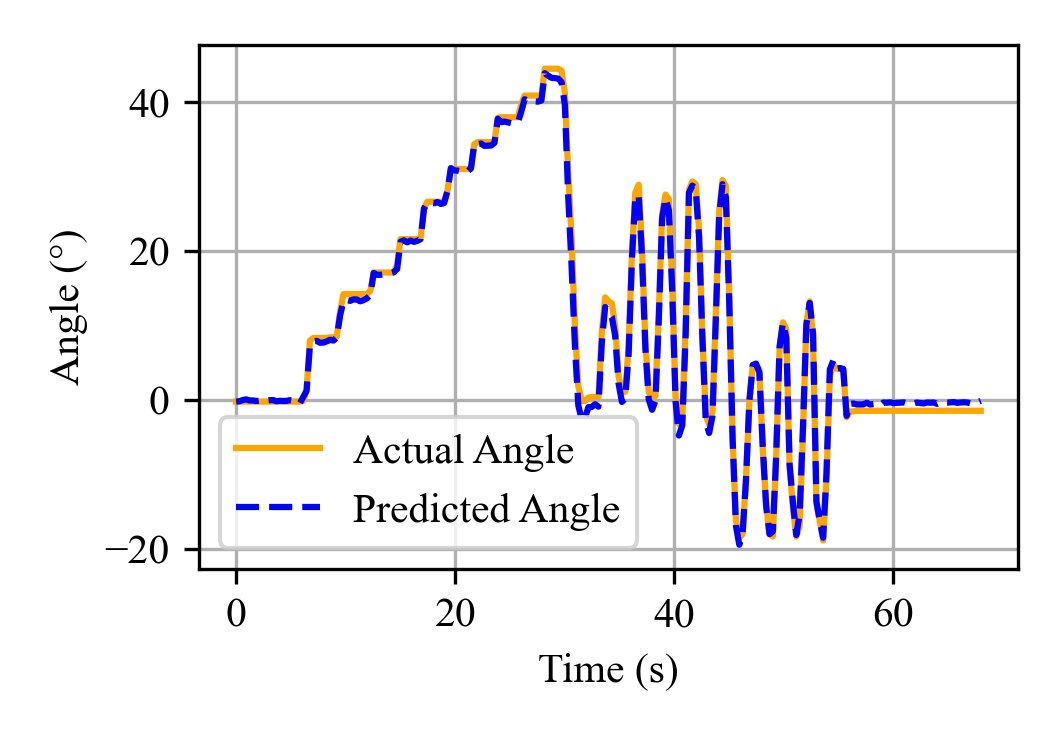}
	\caption{Bending angle prediction validation.} \label{fig:angle_time}
\end{figure}

Figure \ref{fig:angle_time} presents the real-time prediction results obtained during a continuous bending sequence. The predicted angles demonstrate excellent tracking of the actual servo angles, maintaining accuracy even during dynamic transitions. A small shift was observed after bending cycles, primarily due to mechanical effects in the actuation setup, such as slight misalignment between the servo and the plate. Before each run, we reset the servo to the neutral pose, allowed the sensor to settle, and subtracted a fresh baseline. Despite the small offset, repeated cycles showed consistent angle tracking and stable performance. This real-time performance is further validated by the correlation analysis, which displays the relationship between predicted and actual angles across the entire test dataset. The regression model achieved a high coefficient of determination (R² = 0.997) with a root mean square error (RMSE) of 0.90°, confirming both the accuracy and robustness of the proposed approach in capturing bending behavior.
The real-time response characteristics of our system during the bending test are demonstrated in the accompanying video, providing visual confirmation of the system's dynamic performance capabilities. 

\subsection{Touch test}
\begin{figure}[htbp] 
\centering 
\includegraphics[width=0.38\textwidth]{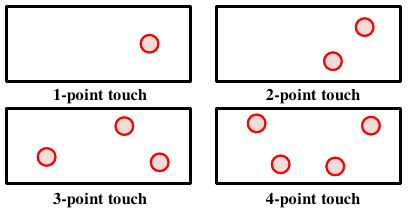} 
\caption{Illustration of multi-point touch locations on the sensor surface.} 
\label{touch_ground_truth} 
\end{figure}

To evaluate the effectiveness of the proposed dynamic adaptive reference strategy, we conducted comprehensive touch mapping experiments under various bending conditions. The experimental setup employed the previously described bending apparatus to control precise angles, while touch stimuli were applied using a cylindrical indenter with a 15 mm diameter. Figure \ref{touch_ground_truth} presents the ground truth locations for one- to four-point touches on the sensor surface. Up to four contact points were used to ensure spatial separability of the touch regions.

\begin{figure}[htbp] 
\centering 
\includegraphics[width=0.48\textwidth]{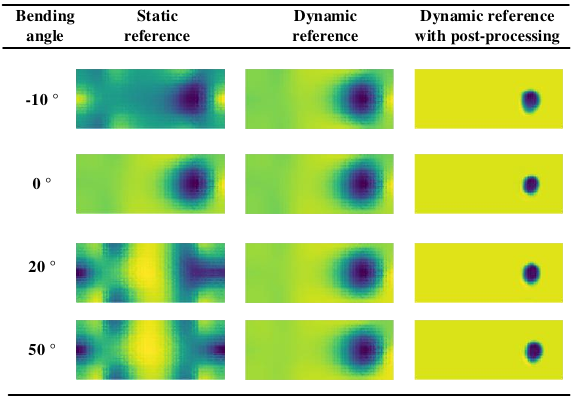} 
\caption{Single point touch mapping under different bending angles with original and dynamic adaptive reference methods.} 
\label{touch_differnt_angle} 
\end{figure}

The performance of our dynamic reference update method was systematically compared to conventional reconstruction approaches lacking reference adaptation.
Figure \ref{touch_differnt_angle} presents single-point touch reconstruction results at various bending angles for three methods: (i) static reference (left column), which does not update the reference during bending; (ii) dynamic reference (middle column), the proposed approach that updates the reference continuously as bending occurs; and (iii) dynamic reference with post‑processing (right column), which applies an additional filtering step to the dynamic‑reference reconstructions to enhance spatial clarity for potential multi‑touch scenarios further. With the static reference method, touch localization is partially maintained at small bending angles (e.g., -10° and 0°) but suffers from reduced clarity and precision. As bending exceeds 20°, reconstruction quality deteriorates markedly, with indistinct localization and strong background artifacts. The dynamic reference method mitigates these effects, preserving clearer and more accurate localization across all bending states. The addition of post‑processing yields sharply defined, artifact‑free contact regions. 
This qualitative improvement is supported by quantitative analysis using the structural similarity index (SSIM): the proposed approach with post‑processing achieves an average SSIM of 0.94, compared to 0.29 for the static reference and 0.71 for the dynamic reference without post‑processing. These results confirm that dynamic reference updating, especially when combined with post‑processing, effectively compensates for bending‑induced reference shifts and preserves spatial fidelity in flexible tactile sensing. 
\begin{figure}[htbp] 
\centering 
\includegraphics[width=0.48\textwidth]{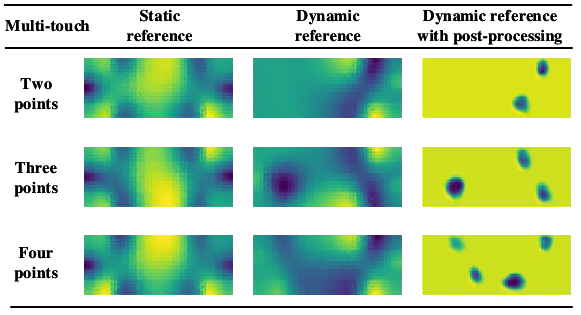} 
\caption{Multi-point touch mapping at 30 degree bending with original and dynamic adaptive reference methods.} 
\label{multi-touchs} 
\end{figure}

To assess the method’s capability under more complex conditions, we tested two-, three-, and four-point contact scenarios (Figure \ref{touch_ground_truth}) at bending angles up to 30°. The corresponding reconstruction results are shown in Figure \ref{multi-touchs}. The static reference method fails to separate individual contacts, producing merged and distorted regions. The dynamic reference approach improves separation but still exhibits residual artifacts. With post‑processing, the proposed method achieves clear and distinct localization of all contact points, as reflected by the highest average SSIM (0.80) compared to the static reference (0.11) and dynamic reference without post‑processing (0.47). In addition, the dynamic reference approach with post-processing enables clear separation and reconstruction of individual contact points, allowing reliable computation of localization error, which was measured as 5.5 mm, 4.3 mm, and 5.0 mm for two, three, and four simultaneous contacts, respectively. This represents a notable advancement in flexible tactile sensing performance.

\subsection{Real‑time demonstration of touch and bending sensing}
\begin{figure}[htbp] 
\centering 
\includegraphics[width=0.48\textwidth]{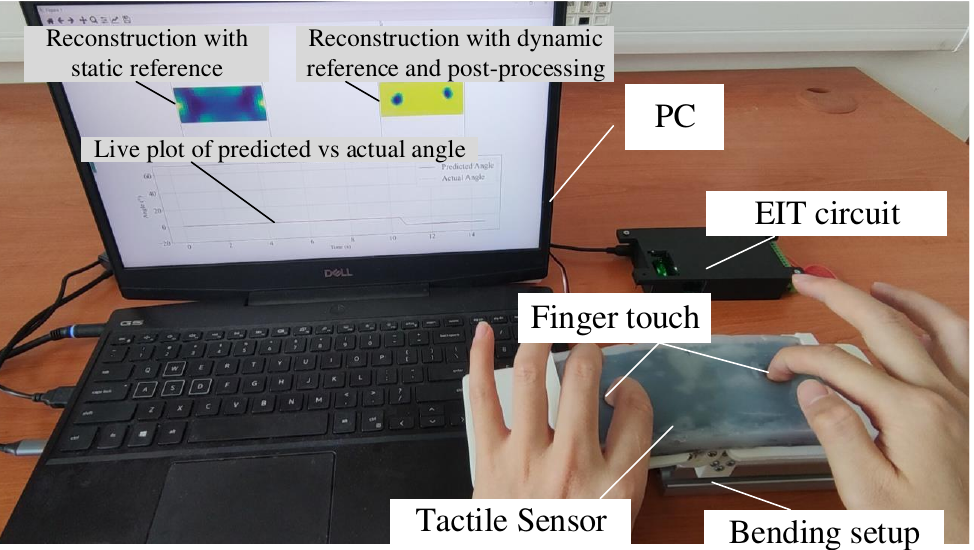} 
\caption{Experimental setup for real-time bend angle prediction and touch detection.} 
\label{demo_setup} 
\end{figure}
To validate the proposed EIT-based sensing framework in a real-world setting, we conducted a real-time experiment involving both bending and multi-point touch interactions. As can be seen in Fig.~\ref{demo_setup}, this setup includes a flexible EIT-based tactile sensor mounted on a motorized bending platform. The EIT circuit acquires impedance measurements, which are transmitted to a PC for real-time processing. The display shows the reconstruction using the static reference alone and the dynamic reference combined with post-processing, and a plot of predicted versus actual bending angles over time. This configuration enables evaluation of both touch and deformation sensing performance.

\begin{figure}[] 
\centering 
\includegraphics[width=0.42\textwidth]{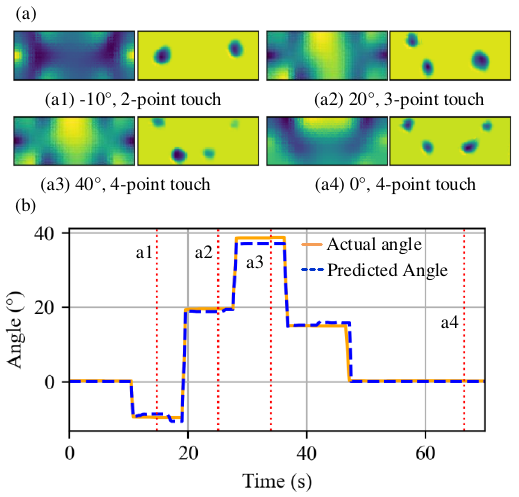} 
\caption{Real-time touch and bending sensing results. } 
\label{demo_touch_bend} 
\end{figure}

As shown in Figure \ref{demo_touch_bend}(a), we compare the touch reconstructions without the proposed framework (using a static reference) against the results produced by our proposed dynamic adaptive reference method.  Four representative time points are selected, corresponding to various bending angles and multi-touch conditions: (a1) -10° with two-point touch, (a2) 20° with three-point touch, (a3) 40° with four-point touch, and (a4) 0° with four-point touch.
At lower bending angles, such as in (a1) and (a4), the static reference method can roughly localize the touch positions, but significant artifacts and shape distortion are still visible. As the bending angle increases (e.g., (a2) and (a3)), these distortions become more severe, and the reconstruction fails to represent the true contact locations, especially under large deformations. In contrast, our proposed framework effectively compensates for bending-induced signal coupling, yielding accurate and artifact-free touch mappings across all configurations. A detailed timing analysis showed an average total processing time of 606.6 ms per frame (1.65 frames $s^-1$), with the biggest latency due to data acquisition (303 ms), state classification (91.6 ms), and reconstruction (86.7 ms). The latency can be further reduced with improved hardware (faster ADC and a dedicated FPGA). Materials with different viscoelastic characteristics and smaller hysteresis may also mitigate the delay.

Figure \ref{demo_touch_bend}(b) further validates the system’s capability to estimate the bending angle and sense multiple touch points in real time. The predicted bending angles closely follow the actual servo motor trajectory, with the key time points (indicated by dashed lines) aligning well with the moments of touch shown in (a).

\section{Conclusion and Future Work}

This work presents a robotic skin that leverages EIT for both touch sensing and bending angle estimation. By introducing a dynamic reference strategy with adaptive signal management, the system effectively handles deformation-induced reference shifts, enabling robust operation under dynamic conditions. A state classification network, enhanced by a linear voltage-to-image embedding, accurately distinguishes between touch, bending, and idle states. For quantitative analysis, a regression model estimates bending angles, while touch locations are mapped through EIT reconstruction followed by lightweight convolutional post-processing to refine spatial resolution. 

The experimental results confirm the sensor’s capability to track continuous bending and resolve multi-touch inputs under various deformation conditions.
Notably, the system achieves high localization precision and bending prediction accuracy, demonstrating the feasibility of the proposed proof-of-concept framework for multimodal robotic skin sensing.

The current prototype is limited by acquisition latency, viscoelastic relaxation of the sensing material, and the non-uniform sensitivity distribution, with higher sensitivity near electrodes and lower sensitivity in the central region. In addition, the present evaluation considers only single-degree-of-freedom bending and fixed indenter geometry. Future work will explore faster acquisition hardware, improved conductive materials with reduced viscoelastic delay, optimized or internal electrode layouts, long-term durability evaluation under repeated loading and prolonged deployment conditions, and data-driven reconstruction methods capable of estimating contact forces, generalizing to arbitrary contact shapes, and extending the approach to multi-axis deformation.

The demonstrated framework holds strong potential for whole-body tactile sensing in humanoid and soft robots, enabling more nuanced physical and social interactions and further advancing the capabilities of robotic skin systems.









\bibliographystyle{IEEEtran}

\bibliography{References}

\end{document}